\documentclass[10pt,twocolumn,letterpaper]{article}

\usepackage{iccv}              
\definecolor{iccvblue}{rgb}{0.21,0.49,0.74}
\usepackage[pagebackref,breaklinks,colorlinks,allcolors=iccvblue]{hyperref}
\definecolor{mygray}{gray}{0.93}
\definecolor{best}{rgb}{0.75,0.90,0.8}
\definecolor{second}{rgb}{0.89,0.93,0.72}
\definecolor{third}{rgb}{0.99,0.97,0.76}
\usepackage{colortbl}
\usepackage{multirow}
\usepackage{rotating}
\usepackage{soul}

\usepackage[accsupp]{axessibility}

\def\confName{ICCV}
\def\confYear{2025}

\title{Domain-aware Category-level Geometry Learning Segmentation for 

3D Point Clouds}

\author{Pei He,    
Lingling Li$^{\ast}$, 
Licheng Jiao\thanks{Corresponding authors} ,
Ronghua Shang,
Fang Liu,
Shuang Wang,
Xu Liu,
Wenping Ma
\\
Xidian University, China\\
{\tt\small hepei@stu.xidian.edu.cn,
llli@xidian.edu.cn, 
lchjiao@mail.xidian.edu.cn}
}

\begin{document}
\maketitle

\begin{abstract}
Domain generalization in 3D segmentation is a critical challenge in deploying models to unseen environments. Current methods mitigate the domain shift by augmenting the data distribution of point clouds. However, the model learns global geometric patterns in point clouds while ignoring the category-level distribution and alignment. In this paper, a category-level geometry learning framework is proposed to explore the domain-invariant geometric features for domain generalized 3D semantic segmentation. Specifically, Category-level Geometry Embedding (CGE) is proposed to perceive the fine-grained geometric properties of point cloud features, which constructs the geometric properties of each class and couples geometric embedding to semantic learning. Secondly, Geometric Consistent Learning (GCL) is proposed to simulate the latent 3D distribution and align the category-level geometric embeddings, allowing the model to focus on the geometric invariant information to improve generalization. Experimental results verify the effectiveness of the proposed method, which has very competitive segmentation accuracy compared with the state-of-the-art domain generalized point cloud methods. The code will be available at https://github.com/ChicalH/DCGL.
\end{abstract}    
\section{Introduction}
\label{sec:intro}
Point clouds segmentation is widely used in autonomous driving \cite{Aoran_survey,LiDAR_survey}, medical analysis \cite{liu2023grab,Yang10130343}, and robot navigation \cite{pami_survey,10558790,feng2020novel}. The point cloud data captured by LiDAR can be used to understand the three-dimensional scene through semantic segmentation, which assigns the category to each point cloud \cite{lai2022stratified,zhang2021attan,9737217}. Although deep learning-based segmentation models have made significant progress, existing methods usually assume that training data and test data have the same distribution \cite{tang2022contrastive,10138737,ma2023explore}. In real scenes, complex data collection conditions, different sensors, and environmental changes will affect the generalization and robustness of point cloud semantic segmentation \cite{xiao20233d,Kim_2023_CVPR,10050808}. Currently, unsupervised domain adaptation (UDA) methods \cite{yi2021complete,10403926_pei} can adapt the source domain (training data) to a specific target domain. 
Although UDA shows promising performance on predefined target domains, its effectiveness decreases when applied to unseen domains.

Domain generalization aims to improve the generalization of unknown domain. Data augmentation that simulates the target domain has been shown to be an effective approach \cite{xiao20233d,he2024domain,zhao2024unimix,park2024rethinking}. 
PointDR \cite{xiao20233d} is a pioneering work in this field that proposes a benchmark dataset and a domain randomization technique to improve generalization in the wild. DGUIL \cite{he2024domain} improves data diversity by simulating point uncertainty in domain shift. And some methods consider synthesizing data patterns under adverse weather\cite{zhao2024unimix, park2024rethinking}. UniMix \cite{zhao2024unimix} proposes a general method to simulate adverse weather data and mix in the source domain. LiDARWeather \cite{park2024rethinking} identifies the key factors of severe weather and proposes a jittering strategy to enhance point cloud data.

However, point clouds are presented in 3D physical space, and how to deal with complex geometric information to improve cross-domain generalization has been ignored. The geometric relations of point clouds generally show cross-domain stability.
However, two issues need to be considered when exploring domain-invariant geometric information in point clouds. On the one hand, the model learns global geometric patterns while ignoring the category-level distribution. Geometric information is important for representing 3D point clouds. Although data augmentation brings some geometric perturbations that allow the model to automatically learn geometric patterns in point clouds, this focuses on the global distribution of the domain while the perturbations make the model ignore the geometric patterns within the class.
On the other hand, the alignment of geometric features before and after the perturbation is ignored. Although augmented data methods are conducive to model learning diverse features, they lack special attention to the domain-invariant information in geometric features.

In this work, we propose category-level geometry learning to explore the domain-invariant geometric features for generalized 3D point clouds. For the first problem, Category-level Geometry Embedding (CGE) is proposed to embed the source features of each category into geometric space and learn correlation parameters to couple the geometric embedding to semantic learning. CGE constructs the embedding of each class separately, enabling the model to perceive the fine-grained geometric properties of categories.
To solve the second problem, Geometric Consistent Learning (GCL) is proposed to learn a domain-invariant geometric representation.
GCL randomly simulates the potential 3D scene with matter accumulation and fuzzy recognition, and aligns the category-level geometric embeddings, which encourages the model to focus on generalized geometric features. The framework provides a geometric perspective on point cloud features, which encourages the model to focus on the geometric invariant information within classes and improves generalization.

Our contributions can be summarized as follows:
\begin{itemize}
\item Category-level Geometry Learning framework is proposed to exploit the domain-invariant geometric features of 3D point clouds to improve domain generalization in unknown domains.
\item Category-level Geometry Embedding (CGE) is proposed to perceive the categoric geometric properties of point cloud features, which constructs the geometric properties of each class and couples geometric embedding to semantic learning. 
\item Geometric Consistent Learning (GCL) is proposed to simulate the latent 3D distribution and align the category-level geometric embeddings, allowing the model to focus on the geometric invariant information to improve generalization.
\item Extensive experiments on multiple benchmarks demonstrate the superior performance of the proposed method over the state-of-the-art domain generalized point cloud segmentation.
\end{itemize}

\section{Related work}
\subsection{3D Semantic Segmentation}

3D semantic segmentation aims to classify each point in point cloud data with semantic label \cite{10234713}. 
The challenge lies in the disorder and sparsity of point cloud data, which is different from 2D image segmentation \cite{10316583, he2022manet} that operates on a structured pixel grid. 
Therefore, point cloud segmentation focuses on how to effectively represent 3D information for unstructured data processing.

Existing 3D point cloud segmentation methods can be divided into three categories according to data processing \cite{Aoran_survey,LiDAR_survey,pami_survey}. Point-based methods \cite{qi2017pointnet, qi2017pointnet++} use variant networks such as PointNet to directly process 3D points. However, due to the complexity of processing large-scale point clouds, these methods usually suffer from huge computational costs. Projection-based methods \cite{Kundu2020VirtualMF, Xu2020SqueezeSegV3SC} convert 3D point clouds into 2D representations and use mature image segmentation models. However, the projection process has inevitable information loss, which limits the performance. Voxel-based methods \cite{choy20194d,maturana2015voxnet} discretize 3D point clouds into voxel grids for calculation, achieving a balance between efficiency and segmentation performance, and becoming a practical choice for a wide range of applications.

Despite substantial progress in 3D segmentation, existing approaches still face challenges in handling domain shifts, particularly in autonomous driving scenarios where adverse weather conditions can degrade performance. This highlights the necessity of developing domain generalization techniques to improve the robustness of 3D point cloud segmentation in real-world applications.

\begin{figure*}[!t]
	\centering
     \vspace{-0.6cm}
	\includegraphics[width=17.7cm]{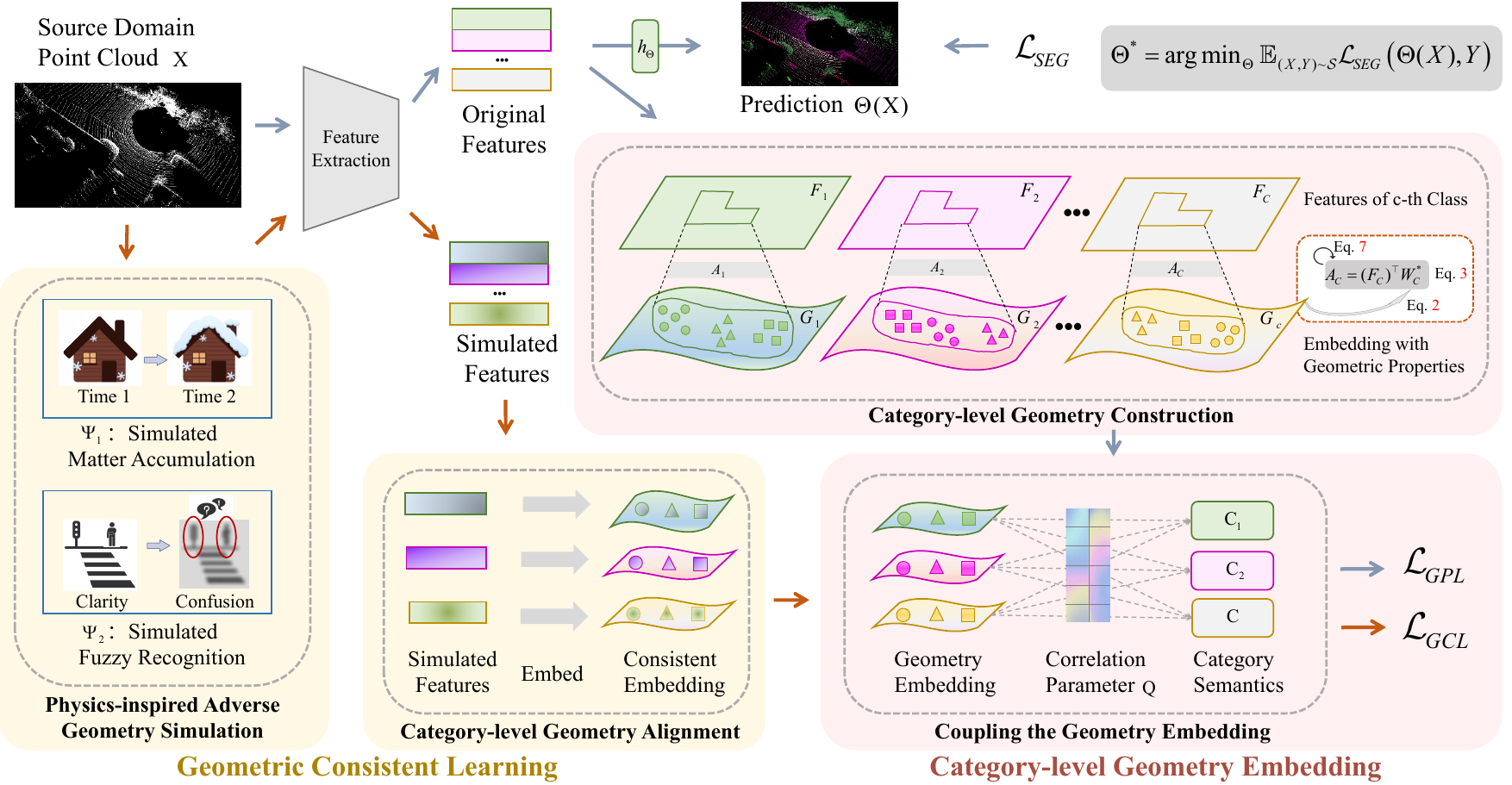}
    \vspace{-0.6cm}
	\caption{\textbf{Overview of the proposed category-level geometry learning framework.} Category-level Geometry Embedding constructs the geometric properties of each class and couples geometric embedding to semantic learning. Geometric Consistent Learning simulates the latent 3D distribution and aligns the geometric embeddings of original and simulated features through consistent learning.}
     \vspace{-0.4cm}
	\label{framwork}
\end{figure*}

\subsection{Domain Generalization for Point Cloud}
Domain generalization (DG) \cite{wang2022generalizing,zhou2022domain} aims to enhance the robustness of model learning so that they can generalize well to unseen target domains without access to target data during training \cite{10144687_pei,10159669_Liao}. Unlike domain adaptation \cite{8661514,10048580} which accesses the target distribution for transfer, DG focuses on learning domain-invariant representations that can generalize across diverse data distributions such as different environments and sensors. 

Data augmentation improves domain generalization by increasing sample diversity, and has been shown to be effective on images \cite{wang2022generalizing,zhou2022domain}. Some point cloud segmentation methods improve model generalization through explicit or implicit data augmentations\cite{xiao20233d,he2024domain,zhao2024unimix,park2024rethinking}. 
PointDR \cite{xiao20233d} proposes a domain randomization technique to enhance point cloud data. DGUIL \cite{he2024domain} proposed domain generalization-aware uncertainty introspective learning to deal with the point uncertainty in domain shift.  UniMix \cite{zhao2024unimix} generates data of different weather conditions and mixes them into the source domain distribution.  
LiDARWeather \cite{park2024rethinking} proposes random jitter and learnable point drop strategies to augment point cloud data.

Although these methods have made significant progress, geometric invariance in 3D point cloud segmentation remains underexplored. The geometric relations of point clouds show the tendency for cross-domain stability: Although the domain shift can significantly affect the distribution of point clouds, the underlying geometric relations of points remain consistent under different acquisition conditions. This paper takes a closer look at the geometric features in point clouds, considering category-level geometric distribution and alignment, which improves the domain generalization of point cloud segmentation by learning domain-invariant geometric representations.

\section{Category-level Geometry Learning}

\subsection{Problem Definition and Framework} \label{Problem}

Given a source domain dataset $\mathcal{S}=\left\{\left(X, Y\right)\right\}_{i=1}^K$, where $X \in  \mathbb{R}^{N \times 4}$ represents a point cloud with $N$ points, each point contains three-dimensional coordinates and intensity information. $Y \in\{1,...,C\}^{N }$ represents the semantic label of each point with $C$ categories. The goal of domain generalization is to learn a segmentation model $\Theta: \mathbb{R}^{N \times 4} \rightarrow\{1, \ldots, C\}^N$, which can achieve high-accuracy point-level classification on any unseen target domain $\mathcal{T}$, using only source domain data $\mathcal{S}$ for training:
\begin{equation}
\Theta^*=\arg \min _\Theta \mathbb{E}_{(X, Y) \sim \mathcal{S}} \mathcal{L}_{SEG}\left(\Theta(X), Y\right),
\end{equation}
where $\mathcal{L}_{SEG}$ is the cross entropy loss for the segmentation. The target domain $\mathcal{T}$ cannot be accessed during training. And the target distribution $P_{\mathcal{T}}(X, Y)$ is inconsistent with the source: $P_{\mathcal{T}}(X, Y) \neq P_{\mathcal{S}}(X, Y)$. Therefore, the model $\Theta$ needs to be robust to the potential domain shift.

To improve generalization, we introduce category-level geometry learning as shown in Figure \ref{framwork}. The source domain point cloud $X$ obtains the original features through the feature extractor, which can be used for classifier $h_{\Theta}$ to predict results. Category-level Geometry Embedding is proposed to map the original features to category-level geometry embedding, which gives the model feedback on the geometric perspective and encourages the model to pay more attention to category-level geometric attributes to predict the results. In addition, Geometric Consistent Learning is proposed to randomly augment the source domain point cloud $X$ and align the category-level geometry embeddings of the original and simulated feature, thereby learning to bridge the domain shift through feedback. This framework focuses on the learning of domain-invariant point cloud geometric features of each category to improve generalization.

\subsection{Category-level Geometry Embedding}\label{modules1}
To achieve category-level geometry embedding, we first construct geometric properties in each class. Then, the geometric embedding is coupled to the model learning process, which enables the model to learn fine-grained geometric representations.

\subsubsection{Category-level Geometry Construction}
Assume that the point cloud is passed through the feature extractor to obtain feature $F \in \mathbb{R}^{N \times D}$, where $N$ represents the number of points, and $D$ is the dimension of point features. We expected a mapping $U^{0}: F \mapsto G$ that can map $F$ with the underlying geometric structure of the point cloud. And it is expected that the mapping is category-unbiased, that is, it contains the geometric properties of all categories. Therefore, the mapping can be represented as $G^{0} = U^{0}(F,G) : \mathbb{R}^{N \times D} \to \mathbb{R}^{N \times C \times M}$, where $C$ is the number of categories and $M$ is the geometric property.

The geometric properties of the feature can be represented in Wasserstein spaces \cite{frogner2018learning}, which are larger than Euclidean spaces to represent metric structures. Therefore, the embedding feature $ G^{0} \in \mathbb{R}^{N \times C \times M}$ is expected to fit the geometric properties of each category in Wasserstein space $W$.
In one batch iteration, the number of points in the $c$-th class is set to $N_{c}$. Therefore, for the $c$-th class, it is expected that $G^{0}_{(c,N_{c})}$ and $W^{0}_{(c,N_{c})}$ are similar, that is, minimize the distance: $min \langle G^{0}_{(c,N_{c})}, W^{0}_{(c,N_{c})}\rangle$ with the constraint that each point has only one unique correspondence and is uniformly distributed in space. This process can be viewed as an instance of the optimal transportation problem \cite{khamis2024scalable} and solved using the Sinkhorn divergence \cite{frogner2018learning} as:
\begin{equation}
	W^*_{(c,N_{c})} =\Delta(u) exp(-\frac{G^{0}_{(c,N_{c})}}{\sigma}) \Delta(v),
	\label{Wc}
\end{equation}
where $\Delta(u)$ and $\Delta(v)$ are diagonal vectors of scaling coefficients, $\sigma$ is the regularization parameter. The optimal geometric feature
$ W^*_{(c,N_{c})}\in \mathbb{R}^{N_{c} \times M}$ includes underlying geometry awareness \cite{khamis2024scalable}, that can take into account some underlying geometric properties of the $c$-th class.

Therefore, for the point feature $F_{(c,N_{c})}$, the mapping from $F_{(c,N_{c})}$ to $ W^*_{(c,N_{c})}$ can be expressed as $U^*(F_{(c,N_{c})},W^*_{(c,N_{c})}): \mathbb{R}^{N_{c} \times D} \to \mathbb{R}^{N_{c} \times M}$.
Assume that this mapping can be expressed as a matrix $ A_{c} \in \mathbb{R}^{D \times M}$, so $U^*(F_{(c,N_{c})},W^*_{(c,N_{c})}) = F_{(c,N_{c})}A_{c}$ and the matrix can be derived by: 
\begin{equation}
A_{c} = ( F_{(c,N_{c})})^\top W^*_{(c,N_{c})}. 
\label{A_c}  
\end{equation}
Concatenating the matrices of all classes to obtain the geometry embedding matrix $A$:
\begin{equation}
A = CON^{C}_{c=1}(( F_{(c,N_{c})})^\top W^*_{(c,N_{c})}),
\label{A}  
\end{equation}
where $A \in \mathbb{R}^{D \times C \times M}$. Therefore, for all points, the category-level geometry embedding can be presented as 
\begin{equation}
U(F,G)_{ \mathbb{R}^{N \times D} \to \mathbb{R}^{N \times C \times M} } = FA.
\label{FA}  
\end{equation}

\subsubsection{Coupling the Geometry Embedding}

The category-level geometry embedding needs to be learned by the model to perceive the geometric property. 
However, this mapping lacks feedback to the model. Therefore, a loss function is required to guide model learning with the correct category semantic information.

For the embedding feature $G=FA$, where $G\in \mathbb{R}^{N \times C \times M}$, each class has $M$ geometric attributes. However, these geometric attributes may not all have a positive effect on point cloud classification. Therefore, adaptive geometric attribute learning can be performed by learning a correlation parameter $Q$.
In addition, the semantic information of the category can be learned by supervision of source label $Y$. Therefore, through backpropagation, the model learns reliable geometric features through the geometry property learning (GPL) loss:
\begin{equation}
\label{category geometry coupling loss}
\mathcal{L}_{GPL}=-\frac{1}{N} \sum_{n=1}^N \sum_{c=1}^C Y_{n c} \log {(||GQ||_{c})}_{n c},
\end{equation}
where $Q\in \mathbb{R}^{ CM \times C}$ represents the relationship matrix between geometric properties and categories, which learns the correlation between geometric embedding and category semantics through adaptive learning. $||\cdot||_{c}$ represents normalization within the category channel, which normalizes all point features to a consistent scale. $\mathcal{L}_{GPL}$ gives the model feedback on the geometric perspective and encourages the model to pay more attention to category-level geometric attributes to predict the results.

Besides, the backpropagation gradient of each iteration should be stable. However, the geometric embedding matrix $A$ is calculated from a batch of data, which is not stable in global training. Therefore, in each iteration, $A$ can be accumulated  through momentum update:
\begin{equation}
	A= \epsilon A+  (1-\epsilon) CON^{C}_{c=1}( (||\dot{A}_{c}||_{2})),
	\label{A_mo}  
\end{equation}
where $\epsilon$ is the momentum coefficient, $||\cdot||_{2}$ represents the L2 normalized of the embedding matrix. 
$\dot{A}_{c}$ further considers enhancing training stability by accumulating only reliable points $\dot{N}_{c}$ in each iteration, which filters true positive points by label and prediction in each category. Then, $\dot{A}_{c}$ can be calculated by $\dot{A}_{c} = (( F_{(c,\dot{N}_{c})})^\top W^*_{(c,\dot{N}_{c})}$.

The proposed CGE first constructs category-level geometric embeddings. Then, the geometric embedding is coupled to the learning process, which enables the model to learn category-aware geometric representations.

\subsection{Geometric Consistent Learning}\label{modules2}
Categorical geometric features are perceived by the network for segmenting point clouds. However, how to further align the category distribution needs to be considered to improve generalization. Therefore, there are two issues to consider. 1. What kind of distribution to align? Because the target domain is unknown in domain generalization, the potential 3D domain shift needs to be effectively simulated. 2. How to align the different domains in a fine-grained manner. For the first issue, we propose a random 3D distribution simulation for the potential domain. For the second issue, we design category-level geometry alignment to align the potential domain with the original domain by consistent learning.

\subsubsection{Physics-inspired Adverse Geometry Simulation}
We first simulate the potential domain shift in 3D scenes. This takes into account that 3D scenes need to conform to more real-world physical regular patterns, which are more complex than flat images. Matter accumulation and confusion recognition need to be taken care of.

In addition, the change of domain is random, so we design a compound random scheme. For the point cloud $X \in  \mathbb{R}^{N \times 4}$, each point $X_{n}$ contains information on three-dimensional coordinates $(x, y, z)$ and reflection intensity $I$. The corresponding label $Y \in  \mathbb{R}^{N}$ represents the semantic category of the point cloud.
We design a random 3D simulation to obtain the augmented sample pair $(X^{\psi},Y^{\psi})$ by processing the information of the original sample:
\begin{equation}
 (X^{\psi},Y^{\psi}) = \sum\nolimits_{i}\eta_{i}  \Psi_{i} (X,Y),  \hspace{0.5em} s.t.\ \eta_{i} \sim B(\beta_{i}),
\end{equation}
where $B$ represents Bernoulli distribution, $\eta_{i} \in \{0,1\} \sim B(\beta_{i})$ means that $\eta_{i} $ gets 1 with a probability of $\beta_{i}$, $\sum_{i}$ means that in one iteration, the simulation $ \Psi_{i}$ are randomly adopted.

\textbf{$\Psi_{1}$: Simulated Matter Accumulation.}
Universal gravitation states that every particle in the universe attracts every other particle with a force, which creates four seasons and scene changes. Typical changes occur when matter in the air accumulates downward, and this accumulation causes changes in the surface of objects. In more specific scenes such as snow and rain, it is not just snowflakes that appear randomly in the air, but more importantly, the accumulation of snow and rain changes the reflective properties of the surface, destroying the inherent characteristics. If there are key surface features for model to recognize covered object, and those features are obscured by external elements, recognition will be hindered. For example, the ground categories such as roads, sidewalks, and terrain, are easily covered, making the boundaries on the ground blurred. For vertically distributed objects such as vegetation and trunks, snow accumulation can change the 3D shape or point cloud features.

Therefore, $\Psi_{1}$ is designed to simulate this matter accumulation. Set the coverage rate to $\rho$, whose physical meaning is the cumulative duration of snowfall or rainfall. 
For the point cloud with a cumulative surface, randomly select $\rho|X|$ object points, where each point has information $(x,y,z,I)$. For the height information $z$, a random accumulation height can be added by $z\prime = z + h$, $s.t.\ h \sim U(h_{1},h_{2})$, where $h$ is a random variable uniformly sampled between $(h_{1},h_{2})$. Meanwhile, due to the influence of the LiDAR reflection signal, the reflection intensity of the point needs to be randomly adjusted by $I\prime = I \cdot \gamma$, $s.t.\ \gamma \sim U(\gamma_{1},\gamma_{2})$, where $\gamma$ is a random variable uniformly sampled between $ (\gamma_{1},\gamma_{2})$. Therefore, the simulated point can be expressed as $(x,y,z\prime,I\prime)$.

\textbf{$\Psi_{2}$: Simulated Fuzzy Recognition.}
LiDAR signals are affected by water droplets and aerosol particles in the air during propagation \cite{zhao2024unimix,hahner2021fog}. Some signals are scattered, causing the returned signal intensity to attenuate. Others are reflected back to the sensor, forming additional backscattering noise. This leads to missing points at long distances and enhancement of false points. Although the transition from clear weather to foggy weather can be achieved by simulating light scattering in a real foggy environment, the generation mode is limited. Inspired by the fuzzy processing in the human brain, people perform contextual reasoning and automatic completion of the absence when their sight is blocked.

Therefore, we design a simple simulation to allow the model to perceive this missing and automatically learn reliable features. We first estimate the echo intensity affected by light scattering \cite{hahner2021fog}. When light scattering makes the echo intensity too low, indicating that the fog is too thick, we set the corresponding label to an unknown category. This slight masking perturbation not only provides a distribution hypothesis, but also allows the model to automatically learn contextual reasoning by more attention to geometric features.

\subsubsection{Category-level Geometry Alignment} 

The enhanced point cloud $X^{\psi}$, together with the original point cloud, is put into the model $\Theta(X^{\psi})$ to obtain the simulated feature $F^{\psi} \in \mathbb{R}^{N \times D}$. Then we use the category-level geometry matrix $A$ obtained from the original distribution to embed $F^{\psi}$ into the geometric space, that is, embed the enhanced feature into the geometry distribution of the original feature and obtain $G^{\psi}=F^{\psi} A$. Therefore, $G^{\psi}\in \mathbb{R}^{N \times C \times M}$ represents the consistent embedding, including $M$ geometric properties for each category.

In addition, consistent learning helps the model learn both the original distribution and the enhanced distribution. Therefore, the correlation parameter $Q$ is also used to learn the relationship between the consistent embedding $G^{\psi}$ and the semantics of the corresponding category $Y^{\psi}$. This can be achieved by calculating the geometric consistency learning (GCL) loss:
\begin{equation}
\label{geometry semantic alignment loss}
\mathcal{L}_{GCL}=-\frac{1}{N} \sum_{n=1}^N \sum_{c=1}^C Y_{n c}^{\psi} \log {(||G^{\psi}Q||_{c})}_{n c}.
\end{equation}

The model performs a backpropagation by accumulating the gradients of $\mathcal{L}_{GCL}$ and $\mathcal{L}_{GPL}$, which are calculated from the original features and the simulated features, respectively. Therefore, the model automatically learns to align the representations of the source domain and the latent domain. The proposed method utilizes category-level geometric learning to push the segmentation model to focus on and learn reliable geometric features to improve generalization.

\section{Experiments and Analysis}

\begin{table*}[!t]
	\centering
    \vspace{-0.6cm}
	\caption{Quantitative comparison between the proposed method and existing generalized point cloud segmentation methods. Top three results are highlighted as
\colorbox{best}{best}, \colorbox{second}{second} and \colorbox{third}{third}, respectively. All models use the MinkUNet18 backbone.}
    \vspace{-0.3cm}
	\label{to_semanticstf}
      \footnotesize
	\renewcommand{\arraystretch}{0.95}
	\setlength{\tabcolsep}{0.55mm}{
        \begin{tabular}{c|cccccccccccccccccccc|cccc|c}
			\toprule
           \toprule
            Methods &\rotatebox{90}{car} & \rotatebox{90}{bi.cle} &\rotatebox{90}{mt.cle} & \rotatebox{90}{truck} & \rotatebox{90}{oth-v.}& \rotatebox{90}{pers.}& \rotatebox{90}{bi.clst}& \rotatebox{90}{mt.clst}& \rotatebox{90}{road}& \rotatebox{90}{parki.}& \rotatebox{90}{sidew.}& \rotatebox{90}{oth-g.}& \rotatebox{90}{build.}& \rotatebox{90}{fence}& \rotatebox{90}{veget.}& \rotatebox{90}{trunk}& \rotatebox{90}{terra.}& \rotatebox{90}{pole}& \rotatebox{90}{traf.}& \rotatebox{90}{mIoU}& \rotatebox{90}{D-fog}& \rotatebox{90}{L-fog}& \rotatebox{90}{Rain}& \rotatebox{90}{Snow}& \rotatebox{90}{Mean}\\
            \midrule 
            \multicolumn{26}{c}{SemanticKITTI $\rightarrow$ SemanticSTF} \\
            \midrule 
PCL \cite{yao2022pcl} &65.9 &0.0 &0.0 &17.7 &0.4 &8.4 &0.0 &0.0 &59.6 &12.0 &35.0 &1.6 &\cellcolor{third}74.0&47.5 &60.7 &15.8 &\cellcolor{third}48.9 &26.1 &27.5 &26.4 &28.9 &27.6 &30.1 &24.6 &27.8\\
MMD \cite{li2018domain} &63.6 &0.0 &2.6 &0.1 &11.4 &28.1 &0.0 &0.0 &\cellcolor{second}67.0 &14.1 &37.9 &0.3 &67.3 &41.2 &57.1 &27.4 &47.9 &28.2 &16.2 &26.9 &30.4 &28.1 &32.8 &25.2 &29.1\\
PointDR \cite{xiao20233d} &67.3 &0.0 &4.5 &19.6 &9.0 &18.8 &2.7 &0.0 &62.6 &12.9 &38.1 &0.6 &73.3 &43.8 &56.4 &32.2 &45.7 &28.7 &27.4 &28.6   &31.3 &29.7 &31.9 &26.2 &29.8  \\	
UniMix \cite{zhao2024unimix} &\cellcolor{third}82.7 &\cellcolor{second}6.6 &8.6 &4.5 &\cellcolor{second}15.1 &35.5 &\cellcolor{best}15.5 &\cellcolor{third}37.7 &55.8 &10.2 &36.2 &1.3 &72.8 &40.1 &49.1 &33.4 &34.9 &23.5 &\cellcolor{third}33.5 &31.4 &34.8 &30.2 &34.9 &30.9 &32.7\\
DGLSS \cite{Kim_2023_CVPR} &72.6 &0.1 &11.7 &\cellcolor{third}29.4 &\cellcolor{third}13.7 &\cellcolor{second}48.3 &0.5 &21.2 &65.0 &\cellcolor{best}20.2 &\cellcolor{third}38.3 &3.8 &\cellcolor{best}78.9 &\cellcolor{second}51.8 &57.0 &\cellcolor{second}36.4 &47.0 &26.9 &34.9  &34.6 &34.2 &\cellcolor{third}34.8  &\cellcolor{third}36.2 &32.1 &34.3\\
DGUIL  \cite{he2024domain}  &77.9 &\cellcolor{best}10.6 &\cellcolor{second}19.1&26.0 &9.7 &\cellcolor{third}46.3 &\cellcolor{third}6.0 &9.3 &\cellcolor{best}69.1 &\cellcolor{second}18.0 &\cellcolor{second}38.6 &\cellcolor{best}9.4 & 73.3 &\cellcolor{third}51.2  &\cellcolor{second}60.8 & 30.9 &\cellcolor{best}50.8 & \cellcolor{best}31.8& 22.3 & \cellcolor{third}35.5 &\cellcolor{second}36.3 &34.5 &35.5 &\cellcolor{second}33.3 &\cellcolor{third}34.8   \\
LiDARWeather \cite{park2024rethinking} &\cellcolor{best}86.1 &\cellcolor{third}4.8 &\cellcolor{third}13.8 &\cellcolor{best}39.7 &\cellcolor{best}26.6 &\cellcolor{best}55.4 &\cellcolor{second}8.5 &\cellcolor{second}50.4 &63.7 &\cellcolor{third}14.9 &37.9 &\cellcolor{third}5.5 &\cellcolor{second}75.2 &\cellcolor{best}52.7 &\cellcolor{third}60.4 &\cellcolor{best}39.7 &44.9 &\cellcolor{third}30.1 &\cellcolor{best}40.8 &\cellcolor{best}39.5 &\cellcolor{third}36.0 &\cellcolor{best}37.5 &\cellcolor{second}37.6 &\cellcolor{third}33.1 &\cellcolor{second}36.1\\
\rowcolor{mygray}	
Ours&\cellcolor{second}84.0 &3.5  &\cellcolor{best}32.8 &\cellcolor{second}37.8 &\cellcolor{third}13.7 &43.9 &5.0 &\cellcolor{best}61.8 &\cellcolor{third}66.8 &11.0 &\cellcolor{best}41.1&\cellcolor{second}8.7 &73.9 &46.6 &\cellcolor{best}61.6 &\cellcolor{third}33.8 &\cellcolor{second}50.1 &\cellcolor{second}31.5 &\cellcolor{second}34.9 &\cellcolor{second}39.1 &\cellcolor{best}39.9 &\cellcolor{second}37.1 &\cellcolor{best}39.5 &\cellcolor{best}33.7 &\cellcolor{best}37.5\\

        \midrule 
        \multicolumn{26}{c}{SynLiDAR $\rightarrow$ SemanticSTF} \\
         \midrule 

MMD \cite{li2018domain} &25.5 &2.3 &2.1 &13.2 &0.7 &22.1 &1.4 &7.5 &30.8 &0.4 &17.6 &0.2 &30.9 &19.7 &37.6 &19.3 &\cellcolor{best}43.5 &9.9 &2.6 &15.1 &17.3 &16.3 &20.0 &12.7 &16.6\\        
PolarMix \cite{xiao2022polarmix}  &39.2 &1.1 &1.2 &8.3 &1.5 &17.8 &0.8 &0.7 &23.3 &1.3 &17.5 &0.4 &45.2 &24.8 &46.2 &20.1 &38.7 &7.6 &1.9 &15.7 &16.1 &15.5 &19.2 &15.6 &16.6\\
PCL \cite{yao2022pcl}  &30.9 &0.8 &1.4 &10.0 &0.4 &23.3 &\cellcolor{second}4.0 &\cellcolor{third}7.9 &28.5 &1.3 &17.7 &\cellcolor{third}1.2 &39.4 &18.5 &40.0 &16.0 &38.6 &12.1 &2.3 &15.5 &17.8 &16.7 &19.3 &14.1 &17.0\\
PointDR \cite{xiao20233d} &37.8 &\cellcolor{third}2.5 &2.4 &\cellcolor{second}23.6 &0.1 &26.3 &2.2 &3.3 &27.9 &7.7 &17.5 &0.5 &47.6 &25.3 &45.7 &21.0 &37.5 &17.9 &5.5  &18.5   &19.5 &19.9 &21.1 &16.9 &19.4 \\	
LiDARWeather \cite{park2024rethinking} &42.1 &\cellcolor{best}2.8 &\cellcolor{second}2.7 &19.2 &0.7 &29.2 &1.9 &4.8 &\cellcolor{third}42.3 &\cellcolor{second}8.7 &\cellcolor{third}21.1 &\cellcolor{second}1.6 &48.2 &\cellcolor{third}26.0 &47.2 &22.1 &32.8 &\cellcolor{third}21.7 &\cellcolor{third}6.5 &20.1 &19.1 &20.7 &22.0 &17.3 &19.8 \\
DGUIL  \cite{he2024domain}  &\cellcolor{third}43.3 &\cellcolor{best}2.8 &\cellcolor{third}2.6 &\cellcolor{third}23.2 &\cellcolor{third}3.2  &\cellcolor{second}31.3 &\cellcolor{third}2.5 &4.4 &34.3 &\cellcolor{best}9.2 &17.9 &0.3 &\cellcolor{second}57.1 &\cellcolor{second}27.6 &\cellcolor{third}50.0 &\cellcolor{second}24.2 &\cellcolor{second}41.5&19.0 &6.1  &\cellcolor{third}21.1 &\cellcolor{third}22.3 &\cellcolor{third}21.9 &\cellcolor{third}26.0 &\cellcolor{third}19.8 &\cellcolor{third}22.5\\
UniMix  \cite{zhao2024unimix}  &\cellcolor{best}65.4 &0.1 &\cellcolor{best}3.9 &16.9 &\cellcolor{second}5.3 &\cellcolor{best}32.3 &2.0 &\cellcolor{second}19.3 &\cellcolor{best}52.1 &5.0 &\cellcolor{best}27.3&\cellcolor{best}3.0 &\cellcolor{third}49.4 &20.3 &\cellcolor{best}58.5 &\cellcolor{third}22.7 &23.2 &\cellcolor{second}26.9 &\cellcolor{second}10.4  &\cellcolor{second}23.4 &\cellcolor{second}24.3 &\cellcolor{second}22.9 & \cellcolor{second}26.1 & \cellcolor{second}20.9 &\cellcolor{second}23.6 \\
\rowcolor{mygray}
Ours &\cellcolor{second}54.4 &\cellcolor{second}2.6 &1.7 &\cellcolor{best}28.0  &\cellcolor{best}5.9 &\cellcolor{third}30.9 &\cellcolor{best}5.6 &\cellcolor{best}23.0 &\cellcolor{second}46.1 &\cellcolor{third}8.0 &\cellcolor{second}21.7 &0.5 &\cellcolor{best}60.3 &\cellcolor{best}29.4 &\cellcolor{second}52.6 &\cellcolor{best}27.5 &\cellcolor{third}40.7 &\cellcolor{best}28.8 &\cellcolor{best}11.0 &\cellcolor{best}25.2 &\cellcolor{best}27.9 &\cellcolor{best}25.1 &\cellcolor{best}28.7 &\cellcolor{best}23.2 &\cellcolor{best}26.3\\

          \midrule 
           \midrule 
	\end{tabular}}
    \vspace{-0.3cm}
\end{table*}

\begin{table*}[!t]
	\centering
	\caption{Comparison of domain generalization on SynLiDAR $\rightarrow$ SemanticKITTI. All models use the MinkUNet18 backbone.}
    \vspace{-0.3cm}
	\label{SynLiDAR_to_kitt}
    \footnotesize
	\renewcommand{\arraystretch}{0.95}
    \setlength{\tabcolsep}{1.1mm}{
        \begin{tabular}{c|ccccccccccccccccccc|c}
			\toprule
           \toprule
            Model  &car & bi.cl &mt.cl & tru. &oth-v. &pers. &bi.cls &mt.cls &road &park. &sidew. &oth-g. &build. &fen. &vege. &tru. &terr. &pole &traf. & mIoU \\
            \midrule 
            PointDR \cite{xiao20233d} &\cellcolor{third}62.0 &\cellcolor{third}6.9 &19.1 &\cellcolor{best}1.8 &6.3 &20.7 &37.4 &4.3 &\cellcolor{second}49.2 &5.3 &\cellcolor{third}36.5 &\cellcolor{best}0.1 &36.3 &\cellcolor{third}12.0 &65.5 &27.9 &\cellcolor{best}45.5 &31.9 &5.5  &24.9 \\
            DGLSS \cite{Kim_2023_CVPR} &\cellcolor{second}62.7 &6.4 &\cellcolor{third}20.4 &\cellcolor{third}1.0 &\cellcolor{third}7.1 &\cellcolor{third}25.4 &\cellcolor{third}39.2 &\cellcolor{third}4.9 &33.7 &\cellcolor{best}7.7 &35.6 &0.0 &\cellcolor{third}44.7 &\cellcolor{second}25.4 &\cellcolor{third}68.0 &\cellcolor{third}35.5 &\cellcolor{second}44.5 &\cellcolor{third}38.3 &\cellcolor{third}10.2  &\cellcolor{third}26.8 \\
            DGUIL  \cite{he2024domain} &61.0 &\cellcolor{second}8.6 &\cellcolor{second}20.5 &0.9 &\cellcolor{best}8.3 &\cellcolor{best}29.9 &\cellcolor{second}43.3 &\cellcolor{second}5.0 &\cellcolor{third}44.1 &\cellcolor{second}7.6 &\cellcolor{second}37.5 &0.0 &\cellcolor{best}56.7 &\cellcolor{best}31.8 &\cellcolor{second}70.1 &\cellcolor{second}36.5 &\cellcolor{third}38.3 &\cellcolor{second}39.0 &\cellcolor{second}14.6 & \cellcolor{second}29.1 \\
            \rowcolor{mygray}
            Ours &\cellcolor{best}74.3 &\cellcolor{best}13.2 &\cellcolor{best}25.8 &\cellcolor{second}1.2 &\cellcolor{second}7.5 &\cellcolor{second}26.9 &\cellcolor{best}58.1 &\cellcolor{best}5.2 &\cellcolor{best}65.2 &\cellcolor{third}6.1 &\cellcolor{best}40.3  &0.0 &\cellcolor{second}56.3 &11.9 &\cellcolor{best}70.5 &\cellcolor{best}38.0 &33.2 &\cellcolor{best}43.7 &\cellcolor{best}17.8  &\cellcolor{best}31.3\\
\midrule 
\midrule 
\end{tabular}}
\vspace{-0.3cm}
\end{table*}

\begin{table*}[!t]
	\centering
	\caption{Domain generalization comparison on MinkUNet34 backbone with SemanticKITTI as source and SemanticSTF as target domain.}
    \vspace{-0.3cm}
	\label{MinkUNet34}
    \footnotesize
	\renewcommand{\arraystretch}{0.95}
    \setlength{\tabcolsep}{1.08mm}{
        \begin{tabular}{c|ccccccccccccccccccc|c}
			\toprule
           \toprule
            Model  &car & bi.cl &mt.cl & tru. &oth-v. &pers. &bi.cls &mt.cls &road &park. &sidew. &oth-g. &build. &fen. &vege. &tru. &terr. &pole &traf. & mIoU \\
            \midrule 
        PointDR \cite{xiao20233d} &\cellcolor{third}72.5 &\cellcolor{second}0.8 &24.5 &28.1 &\cellcolor{best}14.4 &\cellcolor{second}48.6 &\cellcolor{third}4.6 &34.2 &\cellcolor{third}57.4 &\cellcolor{best}16.4 &\cellcolor{third}31.7 &\cellcolor{third}5.2 &72.9 &48.6 &53.9 &\cellcolor{third}35.2 &\cellcolor{third}46.3 &\cellcolor{best}31.2 &24.0  &34.3 \\
        DGLSS \cite{Kim_2023_CVPR} &69.6 &\cellcolor{second}0.8 &\cellcolor{second}42.8 &\cellcolor{best}34.4 &\cellcolor{second}8.9 &\cellcolor{third}41.9 &\cellcolor{best}12.8 &\cellcolor{second}44.5 &52.0 &\cellcolor{third}14.5 &30.8 &\cellcolor{second}6.0 &\cellcolor{third}77.8 &\cellcolor{third}51.1 &\cellcolor{third}57.6 &\cellcolor{best}38.9 &43.2 &\cellcolor{second}29.7 &\cellcolor{third}30.6 &\cellcolor{third}36.2 \\
        DGUIL \cite{he2024domain} &\cellcolor{second}78.2 &\cellcolor{best}2.5 &\cellcolor{third}33.0 &\cellcolor{third}29.7 &\cellcolor{third}6.1 &\cellcolor{best}49.8 &0.8 &\cellcolor{third}40.9 &\cellcolor{best}67.3 &7.2 &\cellcolor{second}38.0 &2.2 &\cellcolor{best}79.8 &\cellcolor{best}54.4 &\cellcolor{best}64.1 &\cellcolor{second}36.8 &\cellcolor{best}52.3 &\cellcolor{third}31.0 &\cellcolor{best}40.0 &\cellcolor{second}37.6 \\
        \rowcolor{mygray}
        Ours &\cellcolor{best}79.5 &\cellcolor{third}0.6 &\cellcolor{best}62.3 &\cellcolor{second}31.4 &4.9 &35.0 &\cellcolor{second}4.8 &\cellcolor{best}63.1 &\cellcolor{second}65.7 &\cellcolor{second}15.8 &\cellcolor{best}39.7 &\cellcolor{best}8.7 &\cellcolor{second}78.6 &\cellcolor{second}52.5 &\cellcolor{second}60.3 &32.0 &\cellcolor{second}50.2 &28.8 &\cellcolor{second}35.4 &\cellcolor{best}39.4 \\
\midrule 
\midrule 
\end{tabular}}
\vspace{-0.4cm}
\end{table*}

\begin{figure*}[!t]
	\centering
    \vspace{-0.8cm}
\includegraphics[width=17.7cm]{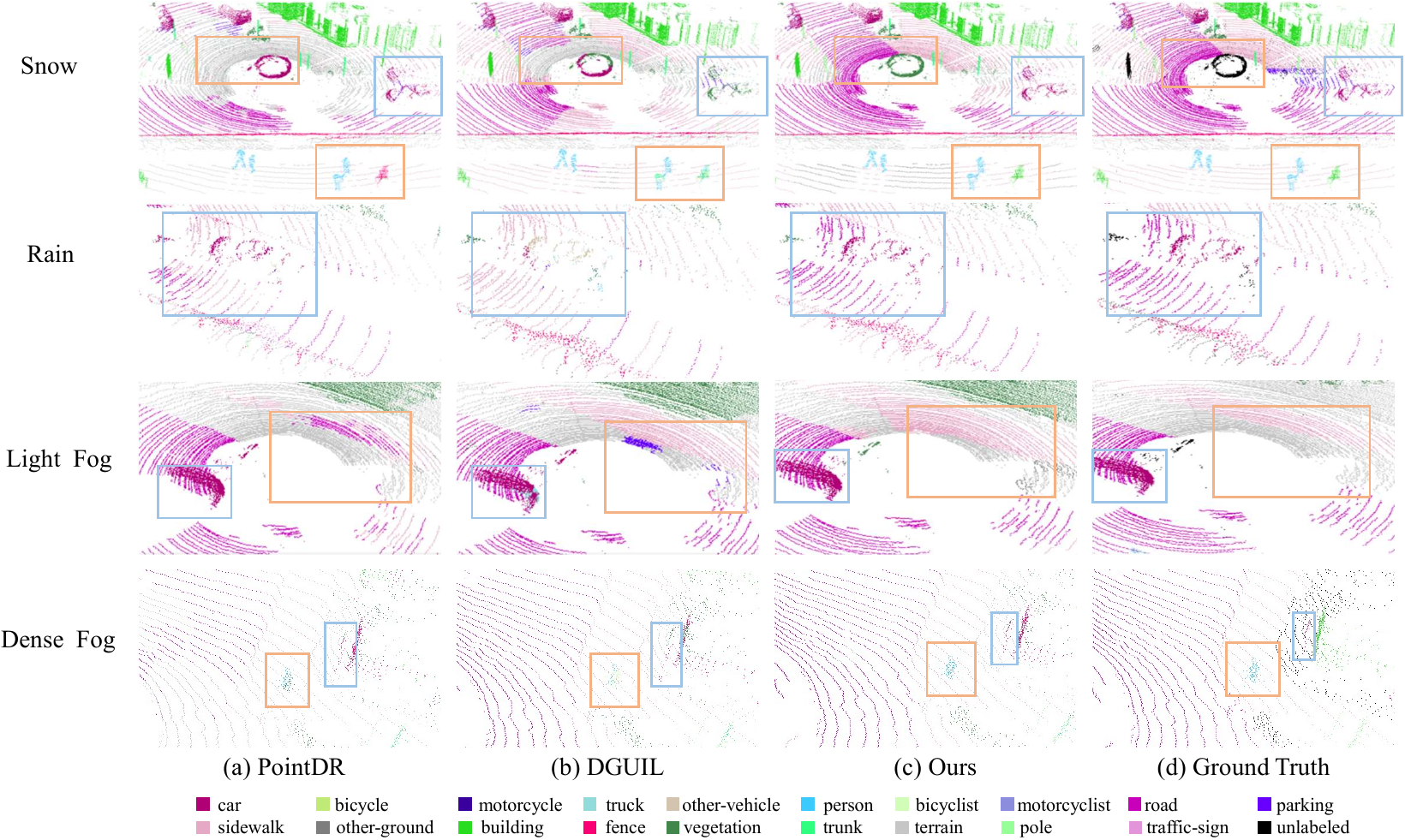}
    \vspace{-0.7cm}
	\caption{Qualitative results of PointDR \cite{xiao20233d}, DGUIL \cite{he2024domain}, and our approach from normal conditions (SemanticKITTI) to adverse weather.}
     \vspace{-0.4cm}
	\label{results}
\end{figure*}

\subsection{Datasets Description}
Three point cloud semantic segmentation datasets are used to compare the methods. 
\textbf{SemanticKITTI} \cite{behley2019semantickitti} consists of LiDAR point clouds captured in urban environments under normal weather conditions. 
The point is annotated with 19 semantic categories, of which the training set serves as the source domain. The validation group is set as the additional target domain.  
\textbf{SynLiDAR} \cite{xiao2022transfer} is a synthetic LiDAR dataset generated from diverse virtual environments, such as downtown, suburban towns, neighborhoods, etc. The training set labeled with 19 categories serves as the source domain.  
\textbf{SemanticSTF} \cite{xiao20233d} comprises LiDAR point clouds of urban scenes collected under adverse weather conditions, such as light fog, dense fog, snow, and rain. SemanticSTF serves as the target domain.

\subsection{Implementation Details}
The framework is built upon PyTorch with an NVIDIA GeForce RTX 3090. For a fair comparison, we follow the previous settings \cite{xiao20233d, zhao2024unimix,he2024domain} that adopt MinkowskiNet \cite{choy20194d} as the segmentation backbone. 
Stochastic gradient descent (SGD) with 0.9 momentum is utilized for optimization. The batch size is 4, and the initialized learning rate is 0.24 with weight decay regularization of 1e-4. 
The training pipeline incorporates multiple random augmentation strategies like previous methods \cite{xiao20233d}, such as affine transformations (full rotation range, scaling in [0.95, 1.05]), horizontal and vertical flipping, jittering, noise injection, and dropout with 0.2 rates, to prevent overfitting of the source domain. The coverage rate $\rho$ is set to 0.3, and $\sigma$ is set to 0.05.

\subsection{Compared with the State-of-the-arts}\label{sota}
We compare the proposal with the state-of-the-art methods on multiple benchmarks, as shown in Tables \ref{to_semanticstf}, \ref{SynLiDAR_to_kitt}, and \ref{MinkUNet34}, illustrating that our method achieves superior performance on all benchmarks.

\textbf{SemanticKITTI $\rightarrow$ SemanticSTF} represents the domain generalization comparison of typical real-world scenarios from normal weather (SemanticKITTI) to adverse weather scenarios (SemanticSTF). The upper part of Table \ref{to_semanticstf} shows the performance comparison from normal to adverse weather, which includes four conditions (Dense fog, Light fog, Rain, and Snow). It can be seen that the proposed method shows stable generalization performance and is superior in most adverse weather conditions, with an average performance of 37.5\% higher than the state-of-the-art method \cite{park2024rethinking} by 1.4\%. It should be noted that generalization in dense fog scenes is advantageous and reaches 39.9\% mIoU, which is 3.9\% higher than LiDARWeather \cite{park2024rethinking}. This is because the model perceives geometric information and can rely on contextual prediction in a noisy environment. In addition, Figure \ref{results} provides an intuitive comparison of the results. Our method shows superior domain generalization capabilities in different target distributions.

\textbf{SynLiDAR $\rightarrow$ SemanticSTF} is the domain generalization from virtual scenes (SynLiDAR) to real adverse weather scenes (SemanticSTF). The lower part of Table \ref{to_semanticstf} shows the comparison results of the proposed method with the outstanding methods. On this benchmark, UniMix \cite{zhao2024unimix} has an advantageous performance of 23.6\% under four realistic adverse weather conditions. The proposed method performs outstandingly, achieving an average performance of 26.3\%. Compared with the state-of-the-art method \cite{zhao2024unimix}, the mIoU in dense fog, light fog, rain, and snow increases by 3.6\%, 2.2\%, 2.6\%, and 2.3\%, respectively. This demonstrates that the proposed geometric learning has a significant advantage in generalization.

\textbf{SynLiDAR $\rightarrow$ SemanticKITTI} represents the generalization from virtual scenes (SynLiDAR) to general real-world scenes (SemanticKITTI). Table \ref{SynLiDAR_to_kitt} shows the results of the proposed method and the current state-of-the-art methods in each category. DGUIL  \cite{he2024domain} achieves an advantageous performance with a mIoU of 29.1\%. The proposed method achieves mIoU of 31.3\%, which is 2.2\% higher than the state-of-the-art method. 

Besides, to demonstrate the integration of the proposed method, the comparison is carried out on another backbone MinkUNet34 \cite{Kim_2023_CVPR}, as shown in Table \ref{MinkUNet34}. The current most advantageous mIoU performance is 37.6\%, while our method achieves 39.4\%. The experiments show that the proposal also has advantages on another backbone.

\vspace{-0.1cm}
\subsection{Ablation Studies}\label{ablation}
\vspace{-0.1cm}
To verify the effectiveness of the proposed modules, we conducted extensive ablation experiments on SemanticKITTI $\rightarrow$ SemanticSTF, as shown in Table \ref{Ablation_Study}. We add the proposed modules one by one to the baseline model, which is trained only with segmentation loss. CGC maps the original features of each category to the geometric space, which can improve mIoU to 32.8\% and increase by 1.4\%. Coupling the category geometry with $\mathcal{L}_{GPL}$ further improves mIoU to 33.0\%. The proposed CGE improves mIoU by 1.6\%, which shows the potential of category-level geometry embedding to improve generalization. GCL significantly boosts the mIoU, among which PAGS reaches 35.3\% mIoU and $\mathcal{L}_{GCL}$ reaches 37.5\%. 
PAGS simulates snow accumulation on object surface and fog recognition, so the generalization is effective in adverse weather. Furthermore, the more fundamental key is the domain-invariant geometry promotes generalization. We verify the impact of removing PAGS on the generalization from virtual to general real world, as shown in Table \ref{PAGS}. The competitive performance further shows that the domain-invariant geometry promotes generalization.
The performance improvement of GCL is based on the model's perception of geometry, focusing on the geometrically invariant information of each class, thereby improving generalization ability. Besides, considering the rotation invariance of point cloud data, TTA is utilized to rotate and scale the point cloud and integrate the prediction results. The results in Table \ref{Ablation_Study} show that TTA can further improve mIoU to 39.1\% based on the proposed modules, which means that the proposal expands the representation space by introducing geometry awareness, giving the prediction a potential advantage.

\begin{table}[!t]
	\centering
    \vspace{-0.7cm}
	\caption{Ablation studies of the modules on SemanticKITTI $\rightarrow$ SemanticSTF. CGE: Category-level Geometry Embedding. CGC: Category-level Geometry Construction. GPL: Geometry Property Learning. GCL: Geometric Consistent Learning. PAGS: Physics-inspired Adverse Geometry Simulation. TTA: Test Time Augmentation.} 
    \vspace{-0.35cm}
\label{Ablation_Study}
\footnotesize
\renewcommand{\arraystretch}{0.95}
\setlength{\tabcolsep}{2.25mm}{
\begin{tabular}{l|cc|cc|cc}
	\midrule
	\multirow{2}{*}{Method} &\multicolumn{2}{c|}{CGE} &\multicolumn{2}{c|}{GCL} &\multirow{2}{*}{mIoU}  & \multirow{2}{*}{$\Delta$}\\
     &CGC &$\mathcal{L}_{GPL}$ &PAGS &$\mathcal{L}_{GCL}$  \\
	\midrule 
	Baseline &-&- &-&-    	&31.4 &-\\
 	 + CGC &\checkmark &- &- &-    &32.8 &+1.4\\
   +$\mathcal{L}_{GPL}$  &\checkmark &\checkmark   &- &-  &33.0 &+1.6\\
	   +PAGS  &\checkmark &\checkmark 	&\checkmark &-	&35.3 &+4.1\\
	   +$\mathcal{L}_{GCL}$  &\checkmark &\checkmark 	&\checkmark &\checkmark	&37.5 &+6.1 \\
    +TTA  &\checkmark &\checkmark 	&\checkmark &\checkmark	&39.1 &+7.7\\
 \midrule 
	\end{tabular}}
\vspace{-0.3cm}
\end{table}

\begin{table}[t!]
	\centering
    \vspace{-0.1cm}
	\caption{Ablations on SynLiDAR $\to$ SemanticKITTI.}
 \vspace{-0.35cm}
\label{PAGS}
 \footnotesize
\renewcommand{\arraystretch}{0.95}
\setlength{\tabcolsep}{4.6mm}{
\begin{tabular}{c|ccccc}
	\midrule
   Method &Baseline &Ours &Ours$^{\ast}$(w/o PAGS) \\
   \midrule
    mIoU &25.1 &31.3  &30.1 \\
 \midrule
\end{tabular}}
\end{table}

\begin{table}[t!]
	\centering
    \vspace{-0.3cm}
	\caption{Ablation studies of geometric awareness and alignment.}
    \footnotesize
    \vspace{-0.3cm}
	\label{geometric}
	\renewcommand{\arraystretch}{0.95}
	\setlength{\tabcolsep}{3.4mm}{
		\begin{tabular}{c|cccc|c}
			\midrule
			  Method &D-fog &L-fog	&rain &snow  &Mean\\
                 \midrule
              Ours  &39.9 &37.1 &39.5 &33.7 &37.5 \\
              w/o geo.  &38.9 &32.6 &37.4 &31.9 &35.2  \\
              w/o ali.  &37.8 &33.2 &33.7 &27.6 &33.1  \\
			\midrule
	\end{tabular}}
    \vspace{-0.4cm}
\end{table}

In addition, to verify the impact of geometric awareness and alignment for adverse weather, additional ablation verification is shown in Table \ref{geometric}. Removing the geometry awareness, i.e., features are not represented in Wasserstein space but in Euclidean space, reduces the performance in all adverse weather conditions. This result illustrates the necessity of geometric feature embedding. Besides, removing the alignment reduces the performance of all conditions, which illustrates the necessity of embedding alignment for stable generalization.

\vspace{-0.1cm}
\subsection{Feature Visualization}\label{Visualization}
\begin{figure}[t]
	\centering
     \vspace{-0.9cm}
	\includegraphics[width=8.5cm]{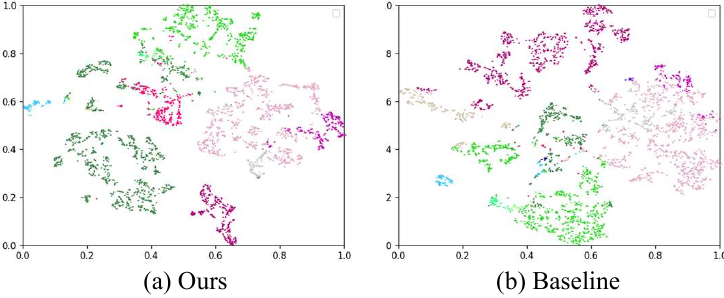}
 \vspace{-0.7cm}
	\caption{Features visualization by t-SNE. The colors and categories have the same corresponding relationship as the results.}
	\label{tsne}
\end{figure}

\begin{figure}[!t]
	\centering
     \vspace{-0.2cm}
	\includegraphics[width=8.5cm]
    {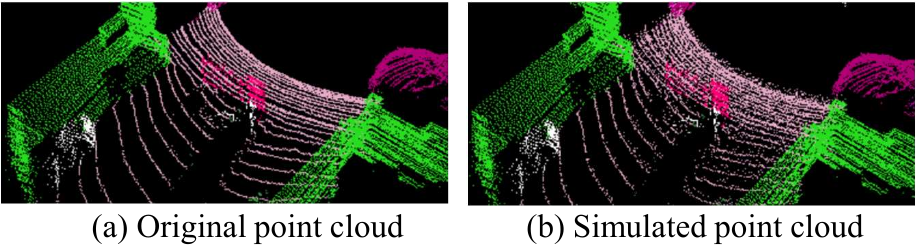}
        \vspace{-0.7cm}
	\caption{Visualization of simulation $\Psi_{1}$.}
	\label{vis_aug}
    \vspace{-0.2cm}
\end{figure}

\vspace{-0.15cm}
In order to intuitively compare the category distribution, we visualized the features using t-SNE \cite{van2008visualizing}, as shown in Figure \ref{tsne}. It can be seen that our method focuses on the distribution of all categories, and the clusters of different categories are slightly separated, which is conducive to better generalization. Besides, Figure \ref{vis_aug} visualizes the local details of the simulation $\Psi_{1}$. Intuitively, the ground points have random height enhancement to simulate the accumulated noise.

\begin{table}[t]
	\centering
	\caption{The effect of $\beta_{1}$ and $\beta_{2}$ on results.}
    \vspace{-0.35cm}
	\label{random}
    \footnotesize
	\renewcommand{\arraystretch}{1.1}
	\setlength{\tabcolsep}{3.9mm}{
\begin{tabular}{c|ccccc}
\hline
    $\beta_{1}$ &0.5 &0.3 &0.3 &0.3 &0.1 \\
    \hline
    $\beta_{2}$ &0.5 &0.5 &0.3 &0.7 &0.5 \\
    \hline
    \rowcolor{mygray}
      mIoU &38.1 &\textbf{39.1} &37.4 &36.8 &36.5 \\
\hline
\end{tabular}}
\vspace{-0.3cm}
\end{table}

\begin{table}[t]
 \vspace{-0.1cm}
  \caption{The effect of different $\epsilon$ on the results.}
  \vspace{-0.35cm}
  \centering
  \label{delta}
  \footnotesize
\renewcommand{\arraystretch}{1.1}
\setlength{\tabcolsep}{3.25mm}{
  \begin{tabular}{cccccc}
   \hline
    $\epsilon$ &0.9 &0.99 &0.999 &0.9999 &0.99999\\
    \hline
    \rowcolor{mygray}
     mIoU  &34.4 &36.9 &39.1 &\textbf{39.4} & 37.0\\
\hline
\end{tabular}}
\vspace{-0.3cm}
\end{table}

\begin{table}[t]
	\centering
	\caption{Hyperparameter ablations of $\rho$, $h_{1}$, $h_{2}$, $ \gamma_{1}$, $\gamma_{2}$.}
 \vspace{-0.35cm}
\label{Hyperparameter}
\footnotesize
\renewcommand{\arraystretch}{1.1}
\setlength{\tabcolsep}{2.8mm}{
\begin{tabular}{c|cccc}
	\hline
	$\rho$  &0.1 &0.2 &0.3 &0.5\\
	\hline
     \rowcolor{mygray}
  mIoU &38.7 &38.9 &\textbf{39.1} &38.6\\
	\hline
    ($h_{1}$,$h_{2}$) &(0.05,0.1)&(0.05,0.3) &(0,0.3)  &(0,0.5)\\
    \hline
     \rowcolor{mygray}
      mIoU &38.4  &\textbf{39.1} &38.7  &38.0 \\
    \hline
    ($\gamma_{1}$, $\gamma_{2}$) &(0.1,1.0) &(0.3,1.0) &(0.5,0.8)  & (0.6,0.7) \\
    \hline
     \rowcolor{mygray}
      mIoU &39.4 &\textbf{39.6} &39.1  &38.0\\
\hline
\end{tabular}}
\vspace{-0.6cm}
\end{table}

\vspace{-0.1cm}
\subsection{Parameter Studies}\label{Parameter}
\vspace{-0.15cm}
In this experiment, we first study the effect of the enhancement ratio of the two simulation methods, as shown in Table \ref{random}. The best results are obtained when probability $\beta_{1}$ is 0.3 and $\beta_{2}$ is 0.5. Secondly, we study the effect of the momentum parameter $\epsilon$ of the matrix A, as shown in Table \ref{delta}. The most favorable performance of \textbf{39.4}\% is achieved when $\epsilon$ is 0.9999, which also shows that is necessary for stable generalization with slow updates. Table \ref{Hyperparameter} provides ablation experiments on hyperparameters $\rho$, $h_{1}$, $h_{2}$, $ \gamma_{1}$, $\gamma_{2}$. The best miou of \textbf{39.6}\% can be achieved when $\rho \hspace{-0.25em} = \hspace{-0.25em}0.3,
\hspace{0.1em}
h_{1}\hspace{-0.25em}=\hspace{-0.25em}0.05,
\hspace{0.1em}
h_{2}\hspace{-0.25em}=\hspace{-0.25em}0.3,
\hspace{0.1em}
\gamma_{1}\hspace{-0.25em}=\hspace{-0.25em}0.3,
\hspace{0.1em}
\gamma_{2}\hspace{-0.25em}=\hspace{-0.25em}1.0$.

\vspace{-0.15cm}
\section{Conclusions} 
\vspace{-0.15cm}
In this paper, a category-level geometry learning framework is proposed to explore the domain-invariant geometric features for domain-generalized 3D point cloud semantic segmentation. Category-level Geometry Embedding (CGE) is proposed to construct the geometric properties of each class and couple geometric embedding to category semantics. Geometric Consistent Learning (GCL) is proposed to simulate the latent 3D distribution and learn domain-invariant geometric representations. Extensive experiments demonstrate that the proposed method achieves state-of-the-art performance on multiple benchmarks. This work bridges a critical research gap in generalized point cloud segmentation. We hope this will inspire future work.

\textbf{Acknowledgments.} This work was supported in part by the Joint Funds of the National Natural Science Foundation of China (U22B2054), the National Natural Science Foundation of China (62076192, 62276199, 62431020 and 62276201), the 111 Project, the Program for Cheung Kong Scholars and Innovative Research Team in University (IRT 15R53), the Science and Technology Innovation Project from the Chinese Ministry of Education, the National Key Laboratory of Human-Machine Hybrid Augmented Intelligence, Xi'an Jiaotong University (HMHAI-202404and HMHAI-202405), the Natural Science Basic Research Program of Shaanxi (Program No.2025JC-YBQN-827).

{
    \small
    \bibliographystyle{ieeenat_fullname}
    \bibliography{main}
}

\end{document}